\documentclass[runningheads]{llncs}
\usepackage{graphicx}
\usepackage{amsmath,amssymb} % define this before the line numbering.
\usepackage{color}

\usepackage{authblk}
\usepackage{cite}
\usepackage{amstext}
\usepackage{colortbl,booktabs}
\usepackage{pfnote}
\begin{document}
\pagestyle{headings}
\mainmatter

\def\ACCV20SubNumber{0408}  % Insert your submission number here

%===========================================================
\title{Synergistic Saliency and Depth Prediction for RGB-D Saliency Detection} % Replace with your title
\titlerunning{Synergistic Saliency and Depth Prediction for RGB-D Saliency Detection}
\author{Yue Wang\inst{1} \and
Yuke Li\inst{2} \and
James H. Elder\inst{3}\and
Runmin Wu\inst{4}\and
Huchuan Lu\inst{1}\thanks{Corresponding author. Email Address: lhchuan@dlut.edu.cn}\and
Lu Zhang\inst{1}}
\authorrunning{Y. Wang et al.}
\institute{Dalian University of Technology\and
UC Berkeley\and
York University \and
The University of Hong Kong}

\maketitle

%===========================================================
\begin{abstract}
Depth information available from an RGB-D camera can be useful in segmenting salient objects when figure/ground cues from RGB channels are weak.  
This has motivated the development of several RGB-D saliency datasets and algorithms that use all four channels of the RGB-D data for both training and inference.  
Unfortunately, existing RGB-D saliency datasets are small, which may lead to overfitting and limited generalization for diverse scenarios.   
Here we propose a semi-supervised system for RGB-D saliency detection
that can be trained on smaller RGB-D saliency datasets {\em without} saliency ground truth,
while also make effective joint use of a large RGB saliency dataset with saliency ground truth together.
To generalize our method on RGB-D saliency datasets, 
a novel prediction-guided cross-refinement module which jointly estimates both saliency and depth 
by mutual refinement between two respective tasks,  
and an adversarial learning approach are employed. 
Critically, our system does not require saliency ground-truth for the RGB-D datasets, 
which saves the massive human labor for hand labeling,
and does not require the depth data for inference, 
allowing the method to be used for the much broader range of applications where only RGB data are available.  
Evaluation on seven RGB-D datasets demonstrates that 
even without saliency ground truth for RGB-D datasets
and using only the RGB data of RGB-D datasets at inference, 
our semi-supervised system performs favorable against state-of-the-art fully-supervised RGB-D saliency detection methods 
that use saliency ground truth for RGB-D datasets at training and depth data at inference
on two largest testing datasets.
Our approach also achieves comparable results on other popular RGB-D saliency benchmarks.
\end{abstract}

%===========================================================
\section{Introduction}

\begin{figure*}[t] 
\centering
\includegraphics[width=0.97\textwidth]{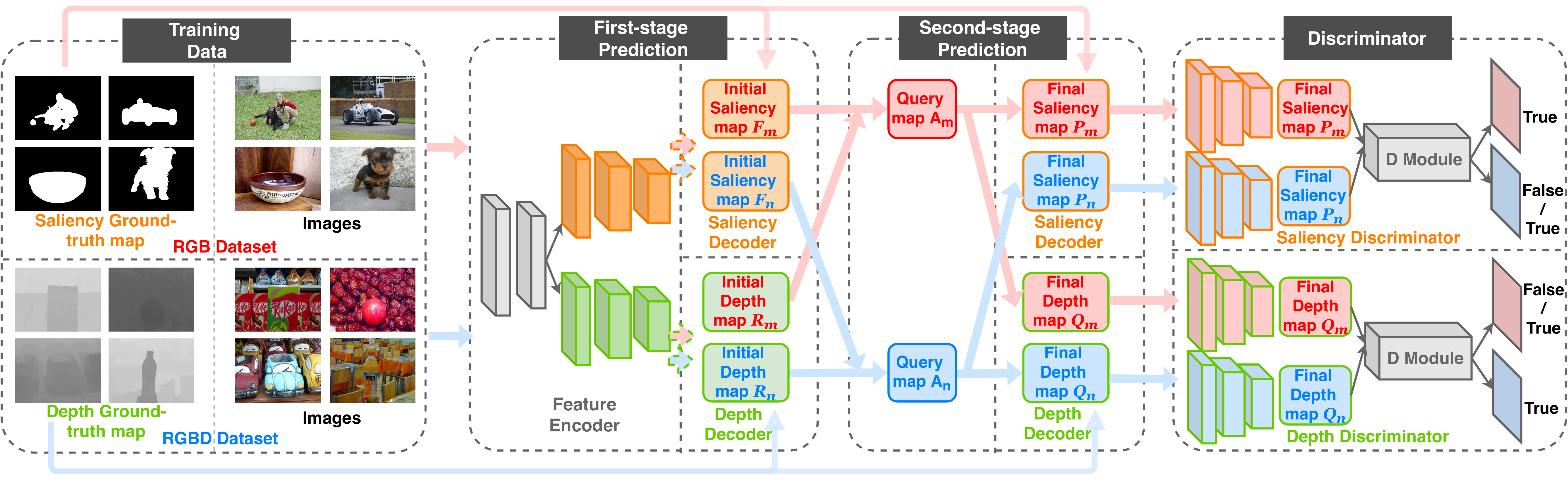}
 \caption{Overview of our proposed method, including the first-stage prediction module, the second-stage prediction module,
 and our discriminator module. The input of our structure is RGB data only, and during testing, we only take RGB data from RGB-D saliency datasets to produce the saliency prediction with the Final Saliency map $P$.}
\label{fig: flow}
\end{figure*}

Salient Object Detection (SOD) aims to accurately segment the main objects in an image at the pixel level.
It is an early vision task important for downstream tasks such as  visual tracking~\cite{lee2018salient}, object detection~\cite{xu2015show}, and image-retrieval~\cite{he2012mobile}.
Recently, deep learning algorithms trained on
large ($>10K$ image) RGB datasets like DUTS~\cite{wang2017learning} have substantially advanced the state of the art. 
However, the problem remains challenging when figure/ground contrast is low or backgrounds are complex. 

It has been observed that in these cases depth information available from an RGB-D camera can be useful in segmenting the salient objects, which are typically in front of the background~\cite{piao2019depth, zhao2019contrast, chen2019three, chen2019multi, chen2018progressively, qu2017rgbd, cong2016saliency, zhu2017innovative}. 
This has motivated the development of several small RGB-D saliency datasets 
\cite{piao2019depth, ju2014depth, peng2014rgbd, li2014saliency, niu2012leveraging, cheng2014depth, fan2019rethinking} with pixel-level hand-labeled saliency ground-truth maps for training.
In order to emphasize the value of depth information, these datasets were constructed so that segmentation based only on RGB channels is difficult due to similarities in colour, texture and 2D configural cues in figure and ground (Fig. \ref{fig: flow}). 
Note that algorithms trained on these datasets use all four channels of the RGB-D data for both training and inference.

Unfortunately, RGB-D images are much rarer than RGB images, and existing RGB-D saliency datasets are much smaller than existing RGB saliency datasets (several hundred vs ten thousand images), which might lead to overfitting and limited generalization for diverse scenarios.  
In theory, one could construct a much larger RGB-D dataset with hand-labeled saliency ground truth, but this would entail specialized equipment and an enormous amount of human labor.
Moreover, the existing fully-supervised methods require the additional depth map as input during inference, 
which limits their applications and makes them be not suitable when only RGB data are available.   

This raises the question: Is it possible to make joint use of large RGB saliency datasets with hand-labeled saliency ground truth,
together with smaller RGB-D saliency datasets {\em without} saliency ground truth, for the problem of saliency detection on RGB-D datasets?  
This would allow us to recruit the massive hand-labeled RGB saliency datasets that already exist 
while facilitating the expansion of RGB-D training datasets, 
since hand-labeled saliency maps for these images is not required.  
Perhaps an even more interesting and ambitious question is: Can we train a semi-supervised system using these two disparate data sources such that it can perform accurate inference on the kinds of images found in RGB-D saliency datasets, {\em even when given only the RGB channels}?  This would allow the system to be used in the much broader range of applications for which only RGB data are available.

However, the images found in RGB-D saliency datasets are statistically different from the images found in typical RGB saliency datasets since they contain more complicated background, 
how to make our semi-supervised model trained with RGB dataset and its saliency ground truth to be generalized well on RGB-D datasets without the saliency ground truth becomes the key challenge.
To address this challenge, we propose our novel prediction-guided cross-refinement network with adversarial learning.

The system consists of three stages (Fig. \ref{fig: flow}).
\textbf{Stage 1.} We build an initial prediction module with two branches: 
a saliency branch that takes RGB images from an RGB saliency dataset as input and is supervised with saliency ground truth; and
a depth branch that takes RGB images from an RGB-D saliency dataset as input and is supervised with depth ground truth (i.e., the Depth data of the RGB-D images).
\textbf{Stage 2.} Since in stage one, for each source dataset, only one branch is supervised,
the statistical difference between two sources makes our initial model not generalize well on the unsupervised source, 
we propose our prediction-guided cross-refinement module as a bridge between two branches.
The supervised branch contributes to the unsupervised one with extra information which promotes the generalization of our model. 
\textbf{Stage 3.} To further solve the distribution difference for two sources in this semi-supervised situation,
we employ a discriminator module trained adversarially, which serves to increase the similarity in representations across sources.

Not only do we train our RGB-D saliency prediction model without the saliency ground truth from RGB-D datasets, 
we also do not need depth data at inference: 
the depth data of RGB-D images is used only as a supervisory signal during training.
This makes our system usable not just for RGB-D data but for the wider range of applications where only RGB data are available.  
We evaluate our approach on seven RGB-D datasets and show that our semi-supervised method achieves comparable performance to 
the state-of-the-art fully-supervised methods, 
which, unlike our approach, use hand-labeled saliency ground truth for RGB-D datasets at training 
and use the depth data at inference.

In summary, we make two main contributions:
\begin{itemize}
\item {We introduce a novel semi-supervised method with prediction-guided cross-refinement module and adversarial learning
that effectively exploits large existing hand-labeled RGB saliency dataset, together with {\em unlabelled
RGB-D data} to accurately predict saliency maps for RGB data from RGB-D saliency datasets.
To the best of our knowledge, our paper is the first exploration of the semi-supervised method for RGB-D saliency detection.
}
\item {We show that, our semi-supervised method
%\footnote{Project Page: https://github.com/wangyue7777/Synergistic-RGB-D-Saliency} 
which does not use saliency ground truth for RGB-D datasets during training and uses only the RGB data at inference, performs favorable against existing fully-supervised methods that use saliency ground truth for RGB-D data at training and use the RGB data as well as depth data at inference on two largest RGB-D testing datasets (SIP and STEREO),
and achieves comparable results on other popular RGB-D saliency benchmarks.}
\end{itemize}

%------------------------------------------------------------------------- 
\section{Related Work}
Considering that it is still a challenge for the existing RGB saliency detections trained on RGB datasets
tend to process images with complex scenarios,
new RGB-D datasets with complex-scenario images and depth data are constructed to focus on
this circumstance
\cite{piao2019depth, ju2014depth, peng2014rgbd, li2014saliency, niu2012leveraging, cheng2014depth}.
The spatial structure information provided by depth data can be of great help for saliency detection,
especially for situations like lower contrast between foreground and background. 
Several methods focus on RGB-D saliency detection have been proposed
to achieve better performance on images with complex scenarios.

In the early stage, approaches like~\cite{peng2014rgbd, cheng2014depth, zhu2017multilayer, cong2016saliency, zhu2017innovative, zhu2017three}
use traditional methods of hand-crafted feature representations, contextual contrast and spatial prior
to extract information and predict saliency maps from both RGB data and depth data in an unsupervised way.
\cite{peng2014rgbd} proposes the first large scale RGB-D benchmark dataset and a detection
algorithm which combines depth information and appearance cues in a coupled manner.
More recently, supervised CNN models that extract high-level content have been found beneficial 
to saliency detection for complex images.
Methods based on CNN structures achieve better performance on RGB-D saliency 
detection~\cite{chen2018progressively, chen2019multi, han2017cnns, qu2017rgbd, zhu2019pdnet, piao2019depth, chen2019three, zhao2019contrast}.
\iffalse
\cite{han2017cnns} employs two CNN network to predict saliency maps from RGB data and depth data, and 
fuse the two networks on prediction level, 
while~\cite{zhu2019pdnet} builds two CNN networks to extract feature from RGB data and depth data, and 
fuse the two branch on feature-level to predict saliency map.
\fi
\cite{han2017cnns} employs two CNN networks to deal with RGB and depth data separately, and 
fuses the two networks on prediction level to predict the final saliency map, 
while~\cite{zhu2019pdnet} fuses the two networks on feature level to predict the final saliency map.
%\cite{zhao2019contrast} enhances the depth clues with a novel contrast-enhanced net to further combine it to feature
%representations from RGB data.
\cite{piao2019depth} applies the multi-scale recurrent attention network to combine features from RGB and depth data
at multiple scales, which considers both global and local information.

However, the above methods suffer from two problems.
First, RGB-D datasets are rarer and the number of images in the existing RGB-D datasets is much smaller, 
which makes the above methods may tend to be overfitting and perform limitedly for diverse situations.
Build larger RGB-D datasets for training would 
require not only massive labor work on labeling the pixel-level ground-truth saliency maps,
but also special equipment to collect depth data.
Second, the above methods demand depth data in both training and inference processes,
which limits the application of RGB-D saliency detection to images with both RGB data and depth data.
In this paper, we propose our semi-supervised method with prediction-guided cross-refinement module and adversarial learning
to predict saliency maps for RGB-D datasets.
With the help of the existing RGB dataset and its saliency ground truth as well as our designed structure,
we are able to train the saliency prediction model for RGB-D datasets
{\em without accessing to their saliency ground truth}.
Besides, by using depth data as an auxiliary task instead of input,
it allows us to evaluate our model with only RGB data.

%------------------------------------------------------------------------- 
\section{Method}

In this paper, we propose a novel semi-supervised approach for RGB-D saliency detection
by exploiting small RGB-D saliency datasets without saliency ground truth, and large RGB saliency dataset with saliency ground truth.
It contains three stages: a first-stage initial prediction module with two branches for saliency and depth tasks 
where each task is supervised with only one source dataset;
a second-stage prediction-guided cross-refinement module 
which provides a bridge between the two branches for each source;
and a third-stage discriminator module which further aligns representations from two sources.
The overview of our proposed structure is shown in Fig.~\ref{fig: flow}.

\subsection{Prediction Module: The First Stage}  \label{sec: gtfs}

\begin{figure*}[t] 
\centering
\includegraphics[width=1.0\textwidth]{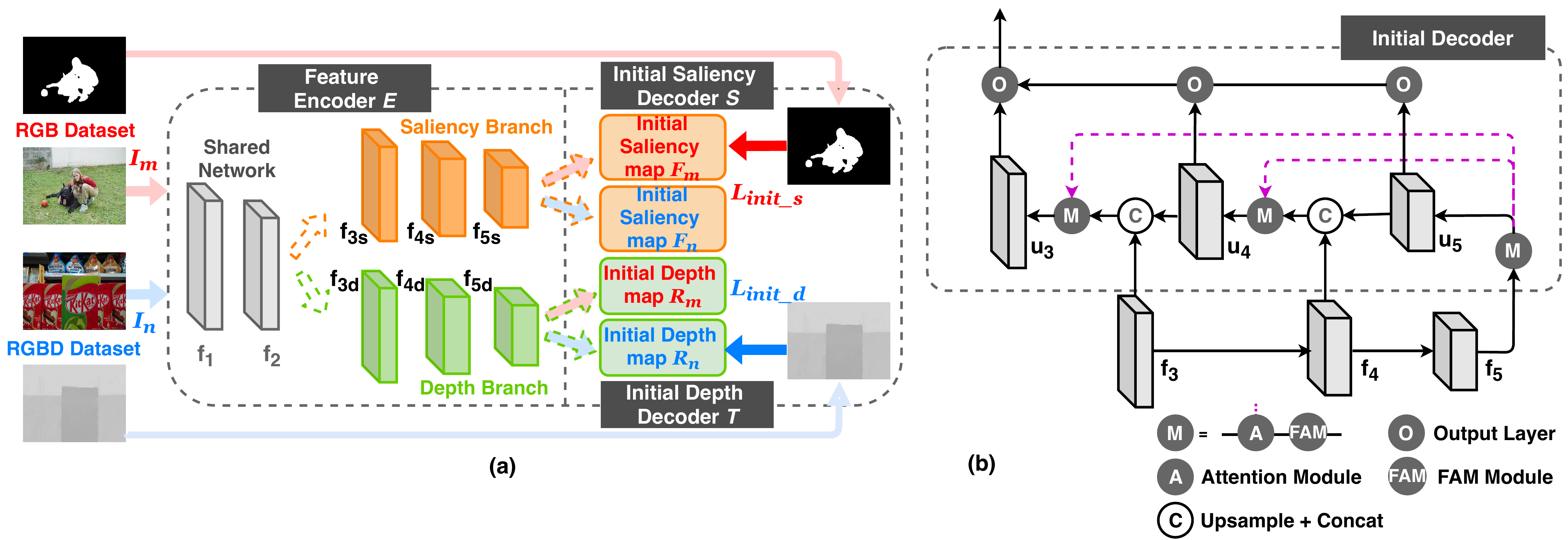}
 \caption{(a) Illustration of our proposed first-stage initial prediction module.
It outputs the initial saliency and depth prediction maps separately for both RGB and RGB-D datasets.
(b) The detailed structure of our decoder, the saliency decoder and depth decoder apply the same structure.}
\label{fig: stage1}
\end{figure*}

The basic structure of our first stage prediction module showed in Fig.~\ref{fig: stage1} consists of a feature encoder $\mathbf E$, 
an initial saliency decoder $\mathbf {S}$, and an initial depth decoder $\mathbf {T}$.
Our $\mathbf E$ is based on a VGG19~\cite{simonyan2014very} backbone, 
which extracts features at five levels.
Both of our decoders $\mathbf {S}$ and $\mathbf {T}$ apply the same architecture 
%which utilizes multi-level features,
using the last three levels of features from the encoder $\mathbf E$, similar to FCN~\cite{shelhamer2017fully}.  
To simultaneously perform two different tasks, 
our $\mathbf E$ is designed to have a two-branch structure for both saliency and depth feature representations.
It has a common two-level root to constrain the model size by sharing weights on the first two levels,
and it is followed by three separate layers for each branch encoding features that are passed to the respective decoders.
In summary, our feature encoder $\mathbf E$ extracts 8 layers of features for each image:
two features layers common to both saliency and depth $ \{f_1, f_2 \}$,
three saliency-specific feature layers $ \{f_{3s}, f_{4s}, f_{5s} \}$,
and three depth-specific feature layers $ \{f_{3d}, f_{4d}, f_{5d} \}$.

To further improve the prediction accuracy,
we incorporate an extra attention module for features on each level for two decoders.
We first introduce a very basic self-attention module from the non-local block~\cite{wang2018non},
which is an implementation of the self-attention form in~\cite{vaswani2017attention}.
Given a query and a key-value pair, the attention function can be described as to learn a weighted sum of
values with the compatibility function of the query and key. 
For self-attention module, query, key, and value are set to be the same,
and the weighted sum output is:
\begin{equation}
\label{atte1}
u = W_z (\text{softmax} ( f^TW_\theta^TW_\phi f) g(f)) + f
\end{equation}  
where $f$ is the input feature, $u$ is the weighted sum output. $W_\theta$, $W_\phi$, $g( \cdot )$ and $W_z$
are the function for query, key, value and weight (See~\cite{vaswani2017attention, wang2018non} for details).

For the highest-level feature $f_5$, we apply the idea of the above self-attention module (Eq.~\ref{atte1}) 
which uses the $f_5$ itself as the query
to obtain the output feature $u_5$.
While for the feature from the other level $f_L$, $L \in \{4, 3\}$, it first need to combine with a higher-level output $u_{L+1}$
by the following common practice:
\begin{equation}
\label{combine1}
\tilde{f}_{L} = conv ({\rm{cat}  (UP } ( u_{L+1}), f_{L}))
\end{equation}  
where $L \in \{3, 4\} $ indicates the level of feature, ${\rm{cat}} (\cdot )$ is the concat function,
${\rm{UP} (\cdot )}$ is the function for upsampling.

We also apply the attention module for the lower-level features.
However, considering the fact that features on different levels are complementary to each other since they extract information
in different resolutions,
high-level features focus on global semantic information, and low-level features provide spatial details which may contain noises,
we would like to select which fine details to pay attention to in low-level features with the global context.
Therefore, the attention module we apply to lower-level features $\tilde{f}_{L}, L \in \{3, 4\}$ 
are based on the highest-level feature $u_5$ to extract meaningful details for prediction.
Based on the idea of Eq.~\ref{atte1}, we replace the query with feature $u_5$ and form our feature-guided attention module.
The overall feature-guided attention module is as follow: 
\begin{equation}
\label{atte2}
u_L=\left\{
\begin{array}{rcl}
W_{z_L} (\text{softmax} ( f_L^TW_{\theta_L}^TW_{\phi_L} f_L) g_L(f_L)) + f_L       &      & {L  =  5}\\
W_{z_L} (\text{softmax} ( u_5^TW_{\theta_L}^TW_{\phi_L} \tilde{f_L}) g_L(\tilde{f_L})) + \tilde{f_L}     &      & { L \in \{4, 3\}}
\end{array} \right. 
\end{equation}
where $\tilde{f_L}$ is the combined feature, and $ u5 $ is the updated feature on $f_5$.

Meanwhile, we also apply the FAM module~\cite{liu2019simple} for all-level features.
It is capable of reducing the aliasing effect of upsampling as well as enlarging the receptive field to improve the performance. 
We then apply three prediction layers on multi-level features $ \{u_5, u_4, u_3 \}$
and add the outputs together to form the initial prediction regarding the branch they belong to.
The detailed architecture of our decoder is illustrated in Fig.~\ref{fig: stage1}(b).

Given an image $ I_m $ from RGB dataset with its saliency ground truth $ Y_m$,
and an image $ I_n $ from RGB-D dataset with its depth data $ Z_n$,
we can obtain their corresponding initial saliency and depth features $ \{u_{3s}, u_{4s}, u_{5s}, u_{3d}, u_{4d}, u_{5d} \}_m $
and $ \{u_{3s}, u_{4s}, u_{5s}, u_{3d}, u_{4d}, u_{5d} \}_n $ with the same encoder $\mathbf E$ 
and separate decoders $\mathbf {S}$, $\mathbf {T}$.
The three levels of saliency features belong to image $ I_m $ will then be used
to output its initial saliency maps $ F_m$,
while the three levels of depth features belong to image $ I_n $ will then be used
to output its initial depth maps $ R_n$.
Since $ Y_m$ of $ I_m $ and $ Z_n$ of $ I_n $ are available,
we can use them to calculate the losses of two initial maps to train our first stage prediction model:
\begin{equation}
\label{salloss1}
\mathcal{L}_{init\_s}(\mathbf E, \mathbf S) = \mathcal{L}_{bce}(F_m, Y_m)
\end{equation}
\begin{equation}
\label{deploss1}
\mathcal{L}_{init\_d}(\mathbf E, \mathbf T) = \mathcal{L}_{1}(R_n, Z_n)
\end{equation}
For the saliency branch, we calculate it using the binary cross-entropy loss $\mathcal{L}_{bce}$,
and for the depth branch, we calculate it using the L1 loss $\mathcal{L}_{1}$.
The overall architecture of our first-stage prediction is illustrated in Fig.~\ref{fig: stage1}.

\begin{figure*}[t] 
\centering
\includegraphics[width=0.98\textwidth]{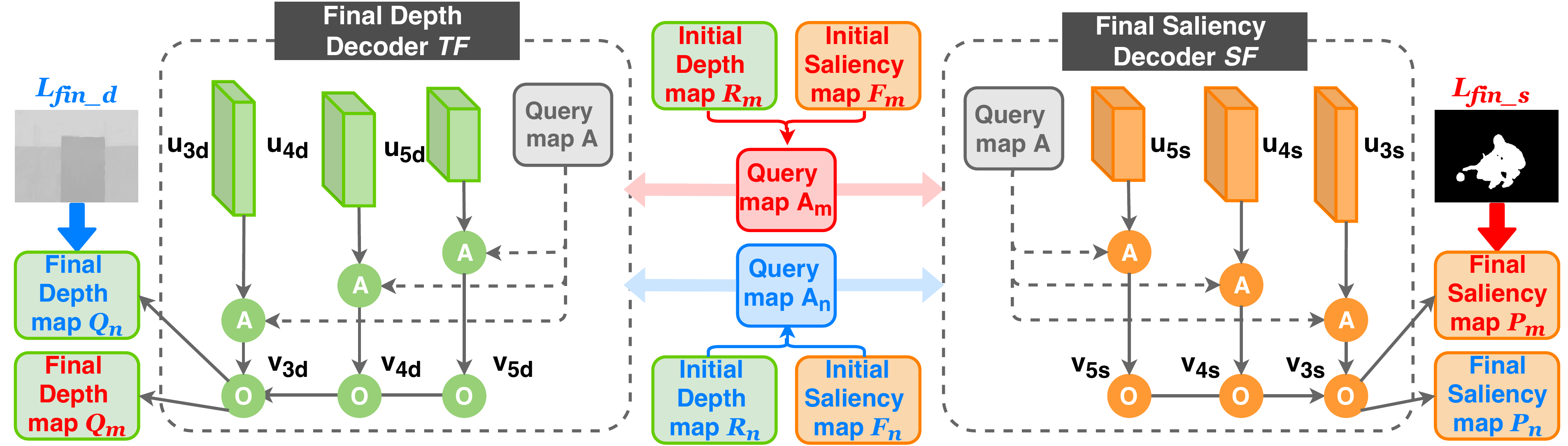}
 \caption{Illustration of our proposed second-stage prediction module.
It uses the initial saliency and depth prediction maps as the query to cross refine feature representations on both branches for RGB dataset and RGB-D dataset.}
\label{fig: stage2}
\end{figure*}

\subsection{Prediction Module: The Second Stage}

In the first stage of our prediction module, 
the saliency and depth branch can only affect each other on the two shared layers in $\mathbf E$. 
Since $\mathbf {S}$ is only supervised with images from the RGB dataset
and $\mathbf {T}$ is only supervised by images from the RGB-D dataset,
these two decoders may not be generalized well on the unsupervised 
source datasets since the difference between RGB and RGB-D saliency datasets on distribution.
However, we notice that the initial depth map $R_n$ from RGB-D dataset which provides spatial structural information 
can be helpful for its saliency prediction,
while the initial saliency map $ F_m$ from RGB dataset can be assisted to its depth prediction
since it shows the location of the important objects which draw people's attention.
Therefore, to enhance the generalization for our initial module on different source datasets, 
we come up with an idea of using a prediction-guided cross-refinement module
as a bridge to transfer information between saliency and depth branch. 

Here, we build our final saliency decoder $\mathbf {SF}$ and final depth decoder $\mathbf {TF}$,
which use our designed prediction-guided method to cross refine the feature representations and initial maps from the first stage.
Our prediction-guided cross-refinement method is based on the same idea of feature-guided attention module in Sec.~\ref{sec: gtfs}.
The detailed structure of our second stage prediction module is shown in Fig.~\ref{fig: stage2}.

In this stage, given features from two branches, 
$ \{u_{3s}, u_{4s}, u_{5s}, u_{3d}, u_{4d}, u_{5d}\}$,
the initial saliency map $ F$ as well as initial depth map $ R $ are used as the query in
the attention module.
We first concat $ F$ and $ R $ to form the query $A$
since the initial $ F$ will also support the saliency branch itself to 
focus on the more informative spatial positions and channels in saliency representations
and it is the same for $R$ to our depth branch.
And then we design a prediction-guided attention module 
with the following equation to update all the multi-level features from two branches.
\begin{equation}
\label{atte3}
v_L = W_{zn_L} (\text{softmax} ( A^TW_{\theta n_{L}}^T W_{\phi n_{L}} u_L) g_{n_L}(u_L)) + u_L 
\end{equation}  
where $u_L$ represents the feature from the first stage and $v_L$ is the updated feature, $ L \in \{5, 4, 3\}$.
All six features from one image $ \{v_{3s}, v_{4s}, v_{5s}, v_{3d}, v_{4d}, v_{5d}\}$
will then be applied to new prediction layers specific to their tasks.
For images from RGB dataset, we sum up three-level saliency outputs to get the final saliency predictions $P_m$,
and then calculate the loss with saliency ground-truth $Y_m$ by:
\begin{equation}
\label{salloss2}
\mathcal{L}_{fin\_s}(\mathbf E, \mathbf {SF}, \mathbf {S,} \mathbf {T}) = \mathcal{L}_{bce} (P_m, Y_m)
\end{equation}

And for images from RGB-D dataset, we also sum up all three-level depth outputs to get the final depth predictions $Q_n$,
and calculate the loss with depth ground-truth $Z_n$ by:
\begin{equation}
\label{deploss2}
\mathcal{L}_{fin\_d}(\mathbf E, \mathbf {TF}, \mathbf {S}, \mathbf {T})= \mathcal{L}_{1} (Q_n, Z_n)
\end{equation}

\subsection{Discriminator}

\begin{figure*}[t] 
\centering
\includegraphics[width=0.98\textwidth]{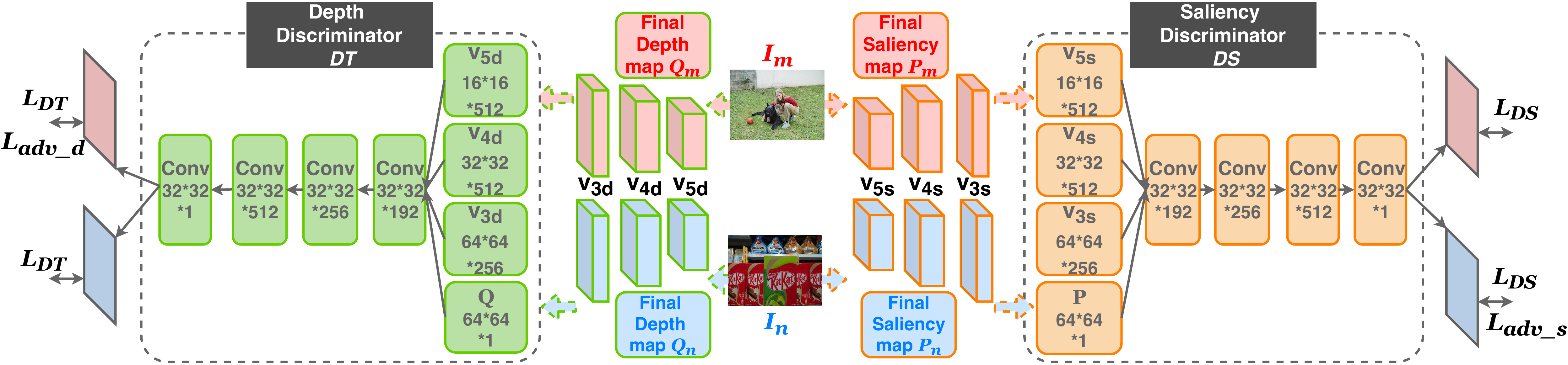}
 \caption{Illustration of our discriminator module for adversarial learning.
It has two parts, the discriminator $DS$ deals with representations from the saliency branch,
and the discriminator $DT$ deals with representations from the depth branch.}
\label{fig: Dim}
\end{figure*}

\begin{figure*}[b] 
\centering
\includegraphics[width=0.95\textwidth]{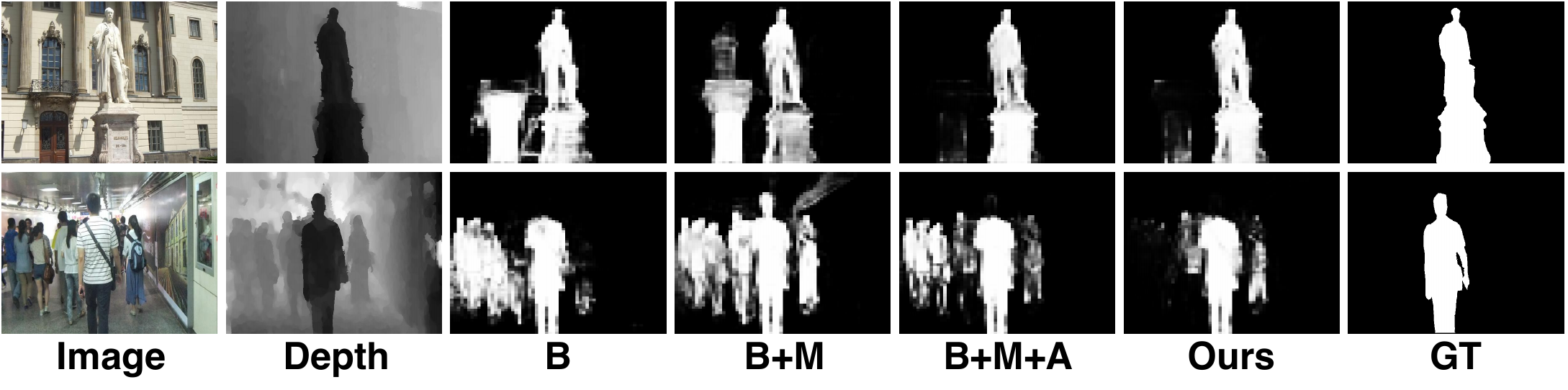}
 \caption{Visual examples for ablation study. See Section \ref{ablation section} for the definition of each subset model.}
\label{fig: vablation}
\end{figure*}

Take the saliency branch as an example, in the first stage, the saliency prediction 
is only supervised by RGB saliency dataset.
Even though we apply our feature-guided attention and FAM module, 
it can be only helpful for improving the performance for saliency detection on RGB saliency dataset itself.
Due to the different distribution between RGB and RGB-D saliency datasets
since images from RGB-D dataset contain more complicated background with similarities to the foreground in color and texture, the improvement on RGB dataset may also cause more noisy background to be detected as the salient region on RGB-D dataset by mistake (Fig.~\ref{fig: vablation} from B to B+M).
With extra depth information from RGB-D saliency dataset and the cross refinement module in our second stage,
the information from the supervised depth branch is able to transfer to the unsupervised saliency branch for RGB-D dataset,
the generalization ability of our saliency prediction model on RGB-D dataset can be enhanced,
such as the first row in Fig.~\ref{fig: vablation} from B+M to B+M+A.
While for some images which are significantly different from RGB dataset on appearance and situations as the one in second row,
the effectiveness of our prediction-guided cross-refinement module may be affected.
To further generalize our saliency model on RGB-D dataset,
we take advantage of the adversarial learning to narrow down the distance 
between the representations from RGB and RGB-D dataset by adding a discriminator module
(Fig.~\ref{fig: vablation} from B+M+A to Ours).
It could also be equally applied to the depth branch,
and the detail of our discriminator module is shown in Fig.~\ref{fig: Dim}.

The original idea of adversarial learning is used for Generative Adversarial Network (GAN)~\cite{goodfellow2014generative},
which is to generate fake images from noise to look real.
It is further used in domain adaptation for
image classification~\cite{hu2018duplex, zhang2019domain}, object detection~\cite{chen2018domain, saito2019strong}
and semantic segmentation~\cite{tsai2018learning, luo2019significance, luo2019taking, vu2019advent}, 
where they train the model on source domain with easily obtained ground truth
and generalize it to target domain without ground truth. 
The purpose of the domain adaptation is to
solve the problem of domain shift due to image difference on appearance, textures, or style for two domains. 
The adversarial learning method uses the generator and discriminator modules to compete against each other 
to minimize the distance between distributions of representations on two domains, 
which is also suitable for our semi-supervised method to further improve its generalization ability.

Our discriminator module has two parts that respond to two task branches,
discriminator $ \mathbf {DS} $ is for the saliency branch,
and discriminator $ \mathbf {DT} $ is for the depth branch.
These two discriminators are trained to distinguish representations from RGB and RGB-D dataset,
and our two-stage prediction module is treated as the generator to fool the discriminators.
The adversarial learning on generator and discriminators helps our prediction model to
extract useful representations for saliency and depth tasks 
which can be generalized on both source datasets.
Here, we align both latent feature representations and output prediction representations
from the two datasets. 
For $\mathbf {DS} $, since image $I_m$ from RGB dataset have the saliency ground truth $Y_m$,
we train $\mathbf {DS}$ so that the saliency feature representations $ \{v_{3s}, v_{4s}, v_{5s}\}_m$ 
and output representation $P_m$ can be classified as source domain label 0,
while the representations $ \{v_{3s}, v_{4s}, v_{5s}\}_n$ and $P_n$ from image $I_n$ 
in RGB-D dataset can be classified as target domain label 1. 
And we calculate the loss of $ \mathbf {DS} $ by:
\begin{equation}
\label{Dloss1}
\begin{aligned}
\mathcal{L}_{DS}(\mathbf {DS})= \ &\mathcal{L}_{bce}(\mathbf {DS}(v_{3s_m}, v_{4s_m}, v_{5s_m}, P_m), 0)
\\ + &\mathcal{L}_{bce}(\mathbf {DS}(v_{3s_n}, v_{4s_n}, v_{5s_n}, P_n), 1)
\end{aligned}
\end{equation}
where $\mathcal{L}_{bce}$ is the binary cross-entropy domain classification loss
since the output channel of our discriminator is 1. 
Meanwhile, instead of predicting one value for the whole image,
we obtain a patch-level output corresponding to the patch-level representations,
which allows the discriminator to predict different labels
for each patch, in order to encourage the system to learn the diversity of factors that determine the generalization for each spatial position.

For $ \mathbf {DT} $, depth representations $ \{v_{3d}, v_{4d}, v_{5d}\}_n$ and $Q_n$ from image $I_n$ are supervised 
by depth ground-truth data $Z_n$,
so we train $\mathbf {DT}$ to classify its representations as source domain label 0,
and classify representations $ \{v_{3d}, v_{4d}, v_{5d}\}_m$ and $Q_m$ from $I_m$ as target domain label 1.
The loss for $ \mathbf {DT} $ is calculated by:
\begin{equation}
\label{Dloss2}
\begin{aligned}
\mathcal{L}_{DT}(\mathbf {DT})= \ &\mathcal{L}_{bce}(\mathbf {DT}(v_{3d_n}, v_{4d_n}, v_{5d_n}, Q_n), 0)
\\ + &\mathcal{L}_{bce}(\mathbf {DT}(v_{3d_m}, v_{4d_m}, v_{5d_m}, Q_m), 1)
\end{aligned}
\end{equation}

To fool $ \mathbf {DS} $, our prediction model is trained to learn saliency representations
$ \{v_{3s}, v_{4s}, v_{5s}\}_n, P_n $ from $I_n$ which can be classified as source domain in $ \mathbf {DS} $.
The adversarial loss for saliency branch can be calculated as:
\begin{equation}
\label{advloss1}
\mathcal{L}_{adv\_s}(\mathbf E, \mathbf {SF}, \mathbf {S}, \mathbf {T}) = \mathcal{L}_{bce}(\mathbf {DS}(v_{3s_n}, v_{4s_n}, v_{5s_n}, P_n), 0)
\end{equation}

For $\mathbf {DT} $, our prediction model is trained to learn depth representations
$ \{v_{3d}, v_{4d}, v_{5d}\}_m, Q_m $ from $I_m$ which can be classified as source domain:
\begin{equation}
\label{advloss2}
\mathcal{L}_{adv\_d}(\mathbf E, \mathbf {TF}, \mathbf {S}, \mathbf {T}) = \mathcal{L}_{bce}(\mathbf {DT}(v_{3d_m}, v_{4d_m}, v_{5d_m}, Q_m), 0)
\end{equation}

\subsection{Complete Training Loss}
\label{sec: ctl}
To summarize, the complete training process includes losses for our prediction model, 
which combines the initial saliency prediction loss for $I_m$ (Eq.~(\ref{salloss1})), 
the initial depth prediction loss for $I_n$ (Eq.~(\ref{deploss1})),
the final saliency prediction loss for $I_m$ (Eq.~(\ref{salloss2})),
the final depth prediction loss for $I_n$ (Eq.~(\ref{deploss2})),
the adversarial loss of saliency branch for $I_n$ (Eq.~(\ref{advloss1})),
the adversarial loss of depth branch for $I_m$ (Eq.~(\ref{advloss2}));
and the losses for saliency and depth discriminators (Eq.~(\ref{Dloss1}), (\ref{Dloss2})), 
%which are the saliency discriminator loss (Eq.~(\ref{Dloss1})),
%and depth discriminator loss (Eq.~(\ref{Dloss2})):
\begin{equation}
\label{total_D}
\mathop{\min}_{\mathbf{DS},  \mathbf{DT}} \mathcal{L}_{DS}  +  \mathcal{L}_{DT} 
\end{equation}
\begin{equation}
\begin{aligned}
\label{total_G}
\mathop{\min}_{\mathbf E, \mathbf {SF}, \mathbf {TF}, \mathbf {S}, \mathbf {T}} \  
& \lambda_{s} \mathcal{L}_{fin\_s} + \lambda_{d} \mathcal{L}_{fin\_d}
\\ + \ \  & \lambda_{init} \lambda_{s}\mathcal{L}_{init\_s} + \lambda_{init} \lambda_{d} \mathcal{L}_{init\_d}
\\ + \ \  & \lambda_{adv\_s} \mathcal{L}_{adv\_s} + \lambda_{adv\_d} \mathcal{L}_{adv\_d} 
\end{aligned}
\end{equation}

%------------------------------------------------------------------------- 
\section{Experiments}

In this section, we evaluate our method and present the experimental results. 
First, we introduce the benchmark datasets and some implementation details of our network architecture. 
Then, we discuss the effectiveness of our method by comparison with the state-of-art methods
and the ablation study.

\subsection{Datasets and Evaluation Metrics}
We evaluate our proposed method on seven widely used RGB-D saliency datasets including
\iffalse
NJUD~\cite{ju2014depth} has 1985 images taken by a Fuji W3 stereo camera;
NLPR~\cite{peng2014rgbd} contains 1000 images constructed by Kinect;
LFSD~\cite{li2014saliency} has 100 images using the Lytro light field camera,
STEREO~\cite{niu2012leveraging} includes 797 images from the Internet,
RGBD135~\cite{cheng2014depth} contains 135 indoor images by Kinect,
SIP~\cite{fan2019rethinking} has 929 images for human activities.
DUT-D~\cite{piao2019depth} contains 1200 images with complex scenes for both indoor and outdoor situations.

NJUD~\cite{ju2014depth} (1985 images),  
NLPR~\cite{peng2014rgbd} (1000 images);
LFSD~\cite{li2014saliency} (100 images),
STEREO~\cite{niu2012leveraging} (797 images),
RGBD135~\cite{cheng2014depth} (135 images),
SIP~\cite{fan2019rethinking} (929 images).
DUT-D~\cite{piao2019depth} (1200 images).
\fi
NJUD~\cite{ju2014depth},  
NLPR~\cite{peng2014rgbd},
LFSD~\cite{li2014saliency},
STEREO~\cite{niu2012leveraging},
RGBD135~\cite{cheng2014depth},
SIP~\cite{fan2019rethinking},
and DUT-D~\cite{piao2019depth}.
To train our model, we use the DUTS~\cite{wang2017learning}, 
an RGB saliency dataset contains 10553 images with saliency ground truth for training the saliency branch. 
And for our depth branch, for a fair comparison, 
we use the selected 1485 NJUD images and 700 NLPR images as in~\cite{piao2019depth}
without saliency ground truth as our RGB-D training set.
We then evaluate our model on 797 images in STEREO, 929 images in SIP 
(These two datasets contain the largest number of images for the test split);
and other testing datasets including 100 images in LFSD, 135 images in RGBD135, 400 images in DUT-D testing set;
as well as the remaining 500 testing images in NJUD, 300 testing images in NLPR.

For quantitative evaluation, we adopt four widely used evaluation metrics
including F-measure($F_m$)~\cite{achanta2009frequency}, mean absolute error (MAE)~\cite{borji2015salient}, 
S-measure($S_m$)~\cite{fan2017structure} and E-measure($E_m$)~\cite{fan2018enhanced}.
In this paper, we report the average F-measure value as $F_m$ which is calculated by the mean of the precision and recall.
\iffalse
For MAE, it calculates the average absolute difference between the prediction and ground-truth.
$S_m$ evaluates the prediction on region-aware and object-aware structural similarity.
And $E_m$ captures global statistics and local pixel matching information.
\fi
For MAE, the lower value indicates the method is better, 
while for all other metrics, the higher value indicates the method is better.

\subsection{Implementation Details}
We apply PyTorch for our implementation using two GeForce RTX 1080 Ti GPU with 22 GB memory. 
For our prediction model, we use VGG19~\cite{simonyan2014very} pre-trained model as the backbone. 
And for the discriminator, we first apply one convolution layer for each input feature/prediction
and concat the latent representations,
then apply four convolution layers to output the one-channel classification result.
\iffalse
Except for the last convolution layer, 
each convolution layer in our discriminator module is followed by a 
Leaky-ReLU~\cite{maas2013rectifier} with a slope of 0.2 for negative inputs.
\fi
We apply ADAM~\cite{kingma2014adam} optimizer for both two-stage prediction module and discriminator module,
with the initial learning rate setting to 1e-4 and 5e-5.
We set $\lambda_{s}=1.75, \lambda_{d}=1.0, \lambda_{init}=0.2, \lambda_{adv\_s}=0.002, \lambda _{adv\_d}=0.001$
to focus more on the saliency branch and the second stage.
All the input images are resized to $256 \times 256$ pixels.

\subsection{Comparison with state-of-the-art methods}
\begin{table*}[t]
\centering\caption{Results on different datasets.
We highlight the best two result in each column in \textcolor{red}{red} and \textcolor{blue}{blue}.}
\centering
\resizebox{0.95\textwidth}{!}{%
\begin{tabular}{c|cccc|cccc|cccc|cccc}
\toprule
 & \multicolumn{4}{c|}{DUT-D} & \multicolumn{4}{c|}{STEREO} & \multicolumn{4}{c|}{SIP} & \multicolumn{4}{c}{RGBD135}\\
 & MAE & $F_m $ & $S_m$ & $E_m$ & MAE & $F_m$ & $S_m$ & $E_m$ & MAE & $F_m$ & $S_m$ & $E_m$ & MAE & $F_m$ & $S_m$ & $E_m$\\
  \hline
DMRA & \textcolor{red}{0.048} & \textcolor{red}{0.883} & \textcolor{red}{0.887} & \textcolor{blue}{0.930} & 
\textcolor{blue}{0.047} & \textcolor{blue}{0.868} & \textcolor{blue}{0.886} & \textcolor{blue}{0.934} 
& 0.088 & 0.815 & 0.800 & 0.858 & 
\textcolor{red}{0.030}& \textcolor{red}{0.867} & \textcolor{red}{0.899} & \textcolor{red}{0.944}\\

CPFP & 0.100 & 0.735 & 0.749 & 0.815 
& 0.054 & 0.827 & 0.871 & 0.902 
& \textcolor{blue}{0.064} & 0.819 & \textcolor{blue}{0.850} & 0.899
& 0.038 & 0.829 & 0.872 & \textcolor{blue}{0.927} \\

TANet & 0.093 & 0.778 & 0.808 & 0.871 & 0.059 & 0.849 & 0.877 & 0.922 
& 0.075 & 0.809 & 0.835 & 0.894 & 0.046 & 0.795 & 0.858 & 0.919 \\

MMCI & 0.112 & 0.753 & 0.791 & 0.856 & 0.080 & 0.812 & 0.856 & 0.894 
& 0.086 & 0.795 & 0.833 & 0.886 & 0.065 & 0.762 & 0.848 & 0.904 \\

PCANet & 0.100 & 0.760 & 0.801 & 0.863 & 0.061 & 0.845 & 0.880 & 0.918 
& 0.071 & \textcolor{blue}{0.825} & 0.842 & \textcolor{blue}{0.900} & 0.050 & 0.774 & 0.843 & 0.912 \\

CTMF & 0.097 & 0.792 & 0.831 & 0.883 & 0.087 & 0.786 & 0.853 & 0.877 
& 0.139 & 0.684 & 0.716 & 0.824 & 0.055 & 0.778 & 0.863 & 0.911 \\

DF & 0.145 & 0.747 & 0.729 & 0.842 & 0.142 & 0.761 & 0.763 & 0.844 
& 0.185 & 0.673 & 0.653 & 0.794 & 0.131 & 0.573 & 0.685 & 0.806 \\
 \hline  
DCMC & 0.243 & 0.405 & 0.499 & 0.712 & 0.150 & 0.762 & 0.745 & 0.838 
& 0.186 & 0.645 & 0.683 & 0.787 & 0.196 & 0.234 & 0.469 & 0.676 \\

CDCP & 0.159 & 0.633 & 0.687 & 0.794 & 0.149 & 0.681 & 0.727 & 0.801 
& 0.224 & 0.495 & 0.595 & 0.722 & 0.120 & 0.594 & 0.709 & 0.810 \\
\hline 

Ours & \textcolor{blue}{0.057} & \textcolor{blue}{0.878}& \textcolor{blue}{0.885} & \textcolor{red}{0.935} 
& \textcolor{red}{0.045} & \textcolor{red}{0.878} & \textcolor{red}{0.893} & \textcolor{red}{0.936} 
 & \textcolor{red}{0.052} & \textcolor{red}{0.856} & \textcolor{red}{0.880} & \textcolor{red}{0.922} 
 & \textcolor{blue}{0.031} & \textcolor{blue}{0.864} & \textcolor{blue}{0.890} & \textcolor{blue}{0.927} \\
 
\bottomrule
\end{tabular}%
}
\label{Tab: table1}
\end{table*}

We compare our method with 9 state-of-the-art RGB-D saliency detection methods including
7 RGB-D deep learning methods:
DMRA~\cite{piao2019depth}, 
CPFP~\cite{zhao2019contrast}, TANet~\cite{chen2019three}, 
MMCI~\cite{chen2019multi}, PCANet~\cite{chen2018progressively}, CTMF~\cite{han2017cnns}, DF~\cite{qu2017rgbd},
and 2 RGB-D traditional methods:
DCMC~\cite{cong2016saliency}, CDCP~\cite{zhu2017innovative}.
The performance of our method compared with the state-of-the-art methods on each evaluation metric is showed
in Table~\ref{Tab: table1} and Table~\ref{Tab: table2}.
For a fair comparison, the saliency maps of the above methods we use are directly provided by authors,
or predicted by their released codes.
We apply the same computation of the evaluation metrics to all the saliency maps. 

For all the listed latest RGB-D methods based on CNNs-based structure, 
they all require depth data as input for both training and inference,
and they use RGB-D saliency ground-truth maps to train the model in a fully-supervised way.
Therefore, they can achieve a good performance on all the datasets.
For RGB-D traditional methods,
they use manually designed cues to calculate the saliency prediction in an unsupervised way,
and they perform worse compared with the CNN-based fully-supervised RGB-D methods.
With the help of images and saliency ground-truth maps from RGB datasets,
our semi-supervised method does not require access to any saliency ground-truth maps for images in RGB-D datasets during training,
and we only require the RGB data without depth data at inference
since we use the depth data as a supervisory signal during training.

\begin{figure*}[t] 
\centering
\includegraphics[width=0.95\textwidth]{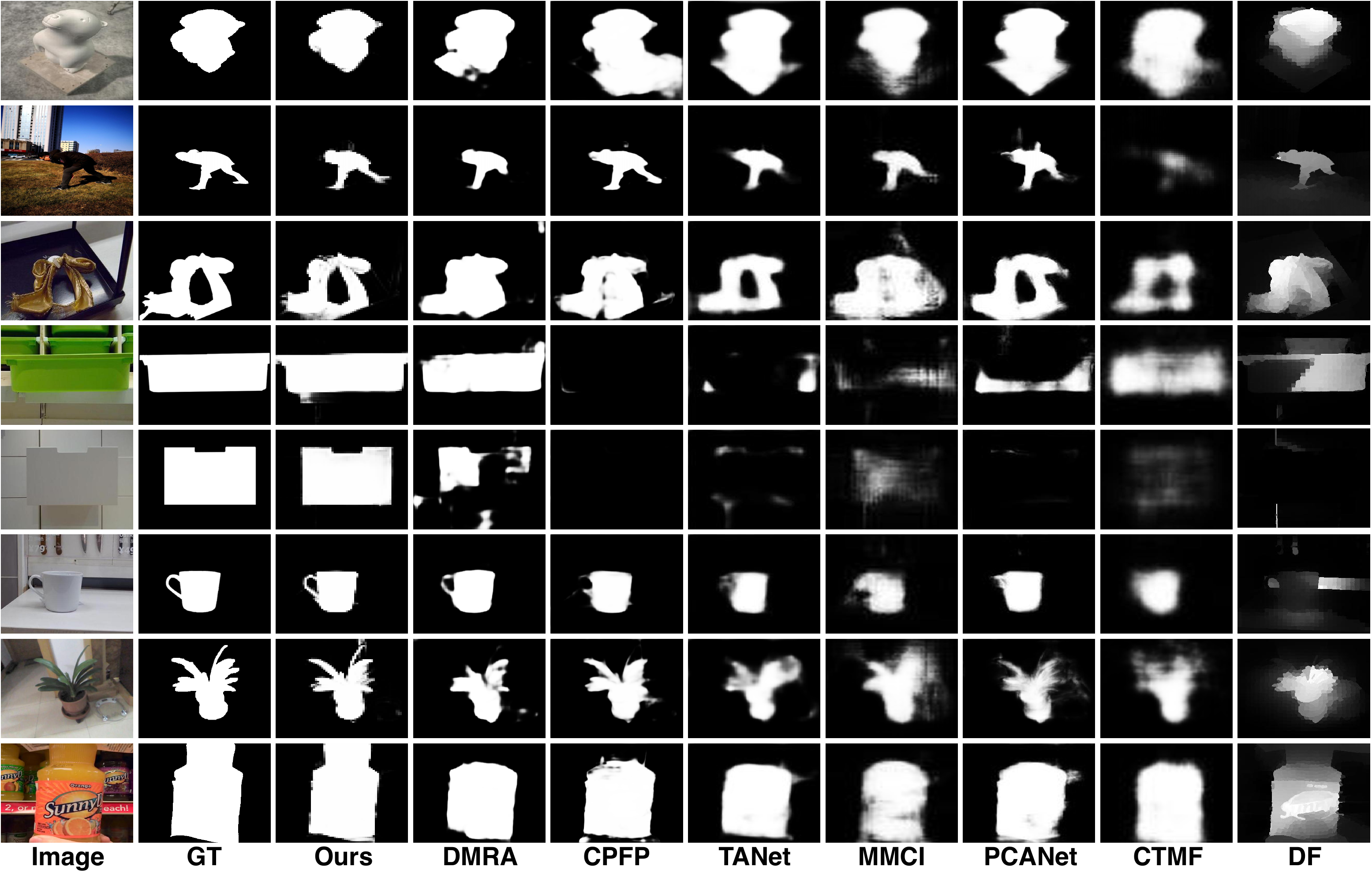}
 \caption{Visual comparison of our method with the state-of-art methods.}
\label{fig: rr}
\end{figure*}

The quantitative results show that, for the two largest testing SIP and STEREO datasets 
containing the largest number of images for testing,  
our semi-supervised method
can achieve better results,
which indicates that our method may generalize better on diverse scenario
even without having access to saliency ground truth for RGB-D datasets.
It can also demonstrate that useful information can be obtained from a larger RGB saliency dataset
and generalized to RGB-D saliency datasets by our designed approach,
despite that the images from these two source datasets have considerable difference
on appearance since RGB-D datasets focus on images with more complicated background.
For other datasets with a smaller number of images for testing such as DUT-D and RGBD135, 
we are also able to reach comparable results with DMRA which are better than other methods.
We may perform slightly worse on two specific datasets, NJUD and NLPR, since 
all other fully-supervised methods use the saliency ground-truth maps from these two datasets during training.
However, we still manage to be comparable with the state-of-art methods on these two datasets.
To better demonstrate the advantage of our method, 
we also present some qualitative saliency examples in Fig.~\ref{fig: rr}. 

\begin{table*}[t]
\centering\caption{Results on different datasets.
We highlight the best two result in each column in \textcolor{red}{red} and \textcolor{blue}{blue}.}
\centering
\resizebox{0.73\textwidth}{!}{%
\begin{tabular}{c|cccc|cccc|cccc}
\toprule
 & \multicolumn{4}{c|}{LFSD} & \multicolumn{4}{c|}{NJUD} & \multicolumn{4}{c}{NLPR}\\
 & MAE & $F_m$ & $S_m$ & $E_m$ & MAE & $F_m$ & $S_m$ & $E_m$ & MAE & $F_m$ & $S_m$ & $E_m$\\
  \hline
DMRA 
& \textcolor{red}{0.076} & \textcolor{red}{0.849} & \textcolor{red}{0.847} & \textcolor{red}{0.899} & 
\textcolor{red}{0.051} & \textcolor{red}{0.872} & \textcolor{red}{0.885} & \textcolor{red}{0.920} & 
\textcolor{red}{0.031}& \textcolor{red}{0.855} & \textcolor{red}{0.898} & \textcolor{red}{0.942}\\

CPFP 
& \textcolor{blue}{0.088} & 0.813 & 0.828 & 0.867 & 
\textcolor{blue}{0.053} & 0.837 & \textcolor{blue}{0.878} & 0.900 
& \textcolor{blue}{0.038} & \textcolor{blue}{0.818} & 0.884 & \textcolor{blue}{0.920} \\

TANet 
& 0.111 & 0.794 & 0.801 & 0.851 & 
0.061 & 0.844 & \textcolor{blue}{0.878} & \textcolor{blue}{0.909} 
& 0.041 & 0.796 & \textcolor{blue}{0.886} & 0.916 \\

MMCI 
& 0.132 & 0.779 & 0.787 & 0.840 & 
0.079 & 0.813 & 0.859 & 0.882 & 0.059 & 0.730 & 0.856 & 0.872 \\

PCANet 
& 0.112 & 0.794 & 0.800 & 0.856 & 
0.059 & 0.844 & 0.877 & \textcolor{blue}{0.909} & 0.044 & 0.795 & 0.874 & 0.916 \\

CTMF 
& 0.120 & 0.781 & 0.796 & 0.851 & 
0.085 & 0.788 & 0.849 & 0.866 & 0.056 & 0.724 & 0.860 & 0.869 \\

DF  
& 0.142 & 0.810 & 0.786 & 0.841 & 
0.151 & 0.744 & 0.735 & 0.818 & 0.100 & 0.683 & 0.769 & 0.840 \\
 \hline  
 
DCMC 
& 0.155 & 0.815 & 0.754 & 0.842 & 
0.167 & 0.715 & 0.703 & 0.796 & 0.196 & 0.328 & 0.550 & 0.685 \\

CDCP 
& 0.199 & 0.634 & 0.658 & 0.737 & 
0.182 & 0.618 & 0.672 & 0.751 & 0.115 & 0.592 & 0.724 & 0.786 \\

 \hline  
 
Ours
 & 0.090 & \textcolor{blue}{0.823} & \textcolor{blue}{0.830} & \textcolor{blue}{0.879} &
 0.055 & \textcolor{blue}{0.852} & \textcolor{blue}{0.878} & \textcolor{blue}{0.909} & 0.044 & 0.809 & 0.875 & 0.915\\

\bottomrule
\end{tabular}%
}
\label{Tab: table2}
\end{table*}

\subsection{Ablation Study}   \label{ablation section}
\begin{table*}[b]
\centering\caption{Ablation Study on our proposed method.
We highlight the best result in each column in \textcolor{red}{red}.}
\centering
\resizebox{0.73\textwidth}{!}{%
\begin{tabular}{l|cccc|cccc|cccc}
\toprule
 & \multicolumn{4}{c|}{NJUD} & \multicolumn{4}{c|}{NLPR} & \multicolumn{4}{c}{STEREO}
 \\
 & MAE & $F_m $ & $S_m$ & $E_m$ & MAE & $F_m$ & $S_m$ & $E_m$ & MAE & $F_m$ & $S_m$ & $E_m$ 
 \\
 \hline
 B & 0.064 & 0.809 & 0.862 & 0.876 
& 0.052 & 0.774 & 0.858 & 0.891 & 
0.053 & 0.835 & 0.877 & 0.908 
\\
 \hline  
B+M & 0.060 & 0.818 & 0.876 & 0.887 
& 0.050 & 0.791 & 0.868 & 0.903 & 
0.053 & 0.854 & 0.889 & 0.921 
\\
 \hline   
B+M+A & \textcolor{red}{0.055} & 0.840 & \textcolor{red}{0.878} & 0.900 
& 0.047 & 0.807 & 0.873 & 0.910 & 
0.050 & 0.868 & 0.888 & 0.928 
\\
 \hline
Ours & \textcolor{red}{0.055} & \textcolor{red}{0.852} & \textcolor{red}{0.878} & \textcolor{red}{0.909} 
& \textcolor{red}{0.044} & \textcolor{red}{0.809} & \textcolor{red}{0.875} & \textcolor{red}{0.915} 
& \textcolor{red}{0.045} & \textcolor{red}{0.878} & \textcolor{red}{0.893} & \textcolor{red}{0.936} 
\\ 
\bottomrule
\end{tabular}%
}
\label{Tab: table3}
\end{table*} 
 
To demonstrate the impact of each component in our overall method,
we conducted our ablation study by evaluating the following subset models:

1) B: Our baseline, a simple saliency detection model directly trained by RGB saliency dataset
with only multi-level fusion in the first stage.

2) B + M: Only trained by RGB saliency dataset while 
adding the FAM module and our feature-guided attention module
in the first stage.

3) B + M + A: Adding the depth branch trained by RGB-D saliency datasets 
and the second stage cross-refinement prediction with the prediction-guided attention module.

4) Ours: Our overall structure with the discriminator module.

Our ablation study is evaluated on three RGB-D datasets and the result is showed in Table~\ref{Tab: table3}.
We also include some visual examples in Fig.~\ref{fig: vablation}. 
It indicates that our baseline model B provides a good initial prediction with the saliency branch trained by the RGB dataset.
By adding feature-guided attention module which helps to focus on more informative spatial positions and channels,
and the FAM module which enlarges the receptive field,
B+M further improves performance by helping saliency detection on RGB saliency dataset.
However, the trained B+M module may not be generalized well on RGB-D saliency detection
due to the different distribution between RGB and RGB-D saliency datasets.
It may perform badly on images with more complicated background (Fig.~\ref{fig: vablation}). 

To improve the generalization ability of our model on RGB-D datasets,
we then add depth branch and the second-stage prediction-guided cross-refinement module
to utilize depth data with spatial structure information for RGB-D images to form B+M+A module.
To further help the generalization of our model on some images from RGB-D datasets 
which have significant difference with images from RGB dataset,
we also add the discriminator module by adversarial learning to align the representations 
on two source datasets for each branch to form our final model.
Table~\ref{Tab: table3} also proves the effectiveness of each module. 

%------------------------------------------------------------------------- 
\section{Conclusions}

In this paper, we propose a novel semi-supervised method for RGB-D saliency detection
with a synergistic saliency and depth prediction way
to deal with the small number of existing RGB-D saliency datasets
without constructing a new dataset.
It allows us to exploit larger existing hand-labeled RGB saliency datasets,
avoid using saliency ground-truth maps from RGB-D datasets during training,
and require only RGB data without depth data at inference. 
The system consists of three stages:
a first-stage initial prediction module to train two separate branches for saliency and depth tasks;
a second-stage prediction-guided cross-refinement module
and a discriminator stage to further improve the generalization on RGB-D dataset
by allowing two branches to provide complementary information and the adversarial learning.
Evaluation on seven RGB-D datasets demonstrates the effectiveness of our method,
by performing favorable against the state-of-art fully-supervised RGB-D saliency methods on two largest RGB-D saliency testing datasets, and achieves comparable results on other popular RGB-D saliency detection benchmarks.

%References are listed in alphabetic order by the surname of the first author, or the identifying word (e.g., in case of a website). Have
%all anonymized references at the beginning of the list.

%here would be your acknowledgement (if any) in the final accepted paper

%===========================================================
\bibliographystyle{splncs}
\bibliography{egbib}

%this would normally be the end of your paper, but you may also have an appendix
%within the given limit of number of pages
\end{document}